\documentclass[11pt]{article}
\usepackage{eamt22}
\usepackage{times}
\usepackage{url}
\usepackage[T1]{fontenc}
\usepackage{latexsym}
\usepackage[small,bf]{caption} 
\usepackage[style=american]{csquotes}
\usepackage{graphicx}
\usepackage{caption}
\usepackage{amsmath}
\usepackage{booktabs}
\usepackage{multirow}
\usepackage{amssymb}
\usepackage{siunitx}
\usepackage{placeins}
\usepackage{subcaption}
\usepackage{enumitem}
\usepackage[dvipsnames]{xcolor}

\title{Controlling Extra-Textual Attributes about Dialogue Participants: \\ A Case Study of English-to-Polish Neural Machine Translation}

\author{Sebastian T. Vincent, Loïc Barrault, Carolina Scarton  \\
  Department of Computer Science, University of Sheffield \\
  Regent Court, 211 Portobello, Sheffield, S1 4DP, UK \\
  \texttt{\{stvincent1, l.barrault, c.scarton\}@shef.ac.uk}}

\date{}

\begin{document}
\maketitle
\begin{abstract}
Unlike English, morphologically rich languages can reveal characteristics of speakers or their conversational partners, such as gender and number, via pronouns, morphological endings of words and syntax. When translating from English to such languages, a machine translation model needs to opt for a certain interpretation of textual context, which may lead to serious translation errors if extra-textual information is unavailable. We investigate this challenge in the English-to-Polish language direction. We focus on the underresearched problem of utilising external metadata in automatic translation of TV dialogue, proposing a case study where a wide range of approaches for controlling attributes in translation is employed in a multi-attribute scenario. The best model achieves an improvement of $+5.81$ chrF++$/{+}6.03$ BLEU, with other models achieving competitive performance. We additionally contribute a novel attribute-annotated dataset of Polish TV dialogue and a morphological analysis script used to evaluate attribute control in models.

\end{abstract}

\section{Introduction}
In some languages, dialogue explicitly expresses certain information about the interlocutors: for example, while in English words describing the speaker \enquote{I} and the interlocutor \enquote{you} are ambiguous w.r.t. their gender, number and formality, languages such as Polish, German or Spanish will mark for one or more of these attributes. In industrial settings such as dubbing and speech translation, there is an abundance of available metadata about the interlocutors, such as their genders, that could be used to help resolve these ambiguities.

\begin{table}[ht]
\centering
\scalebox{.8}{
\begin{tabular}{rl}
\toprule
Field & Value\\
\midrule
source & "Are you blind?" \\
spoken by (=speaker) & "Anne" \\
speaker's gender & "feminine" \\
spoken to (=interlocutor(s)) & ["Mark", "Colin"] \\
interlocutor(s)' gender & "masculine" \\
formality & "informal" \\
\bottomrule
\end{tabular}}
\caption{A TV segment along with available metadata.}
\label{tab:industrial}
\end{table}

Table \ref{tab:industrial} shows an example of such a TV segment: the English sentence \textit{`Are you blind?'}, should translate to Polish as \textit{`Jesteście ślepi?'} as the addressee is a group of men and the setting is informal; however, when spoken e.g. formally to a mixed-gender group of people, the correct translation would read \textit{`Są państwo ślepi?'}, using a different verb inflection and an honorific \textit{państwo}.
Since the contextual information required to resolve the ambiguity in this example does not belong to the text itself, traditional models do not use it. This yields hypotheses which introduce some assumptions about that context, typically reflecting biases present in the (often unbalanced) training data. To avoid this, a better solution is to resolve such ambiguities by using both the available metadata and the source text as translation input. Alternatively, when such information is unavailable, all possible contextual variants could be provided as output, passing the choice from the model to the user~\cite{Jacovi2021trust,schioppa-etal-2021-controlling}.

\begin{figure*}[h]
\centering
\scalebox{.75}{
\includegraphics[width=\linewidth]{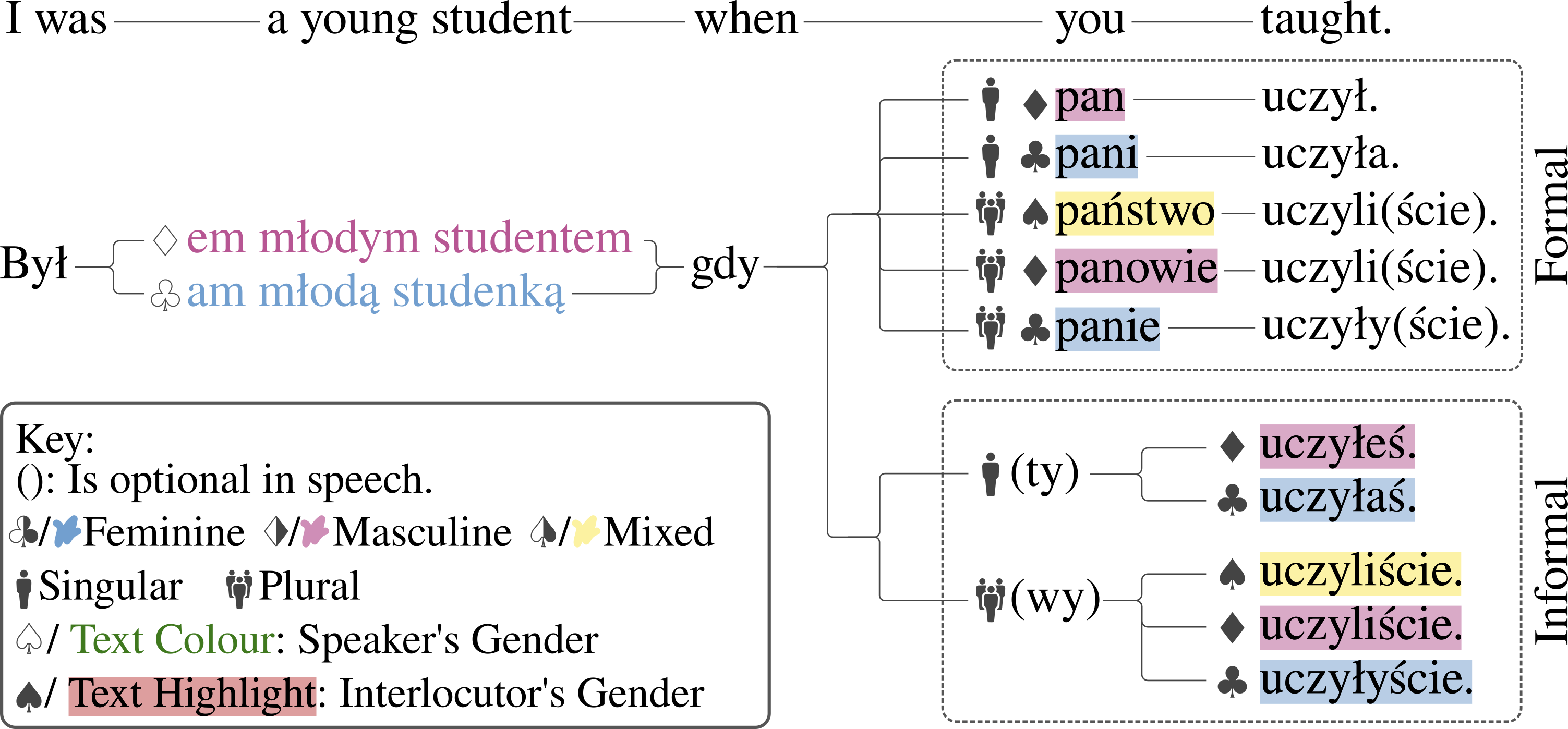}}
\caption{Example of an ambiguous English sentence with all plausible translations to Polish. There are a total of 18 equally plausible possible hypotheses based on the combination of contexts.}

\label{fig:example_coh}
\end{figure*}

In the context of the gender of the speaker and interlocutor, prior research has explored two ways in which such information influences a text~\cite{Rabinovich2017,Vanmassenhove2020}. Firstly, naturally occurring texts satisfy grammatical agreement between the gender of the speaker and interlocutor and the utterances which describe them. How this agreement is expressed in speech varies among different languages~\cite{stahlberg-etal-2007-representation}. Polish is a \textit{grammatical gender language}: every noun is assigned a gender, and grammatical forms must agree with that noun. In contrast, English is a \textit{natural gender language}, with \enquote{no grammatical markings of sex}~\cite[p. 165]{stahlberg-etal-2007-representation}. Secondly, gender can be seen as a demographic factor that influences the way people express themselves (e.g. word choice). Hereinafter we refer to the former as \textit{grammatical agreement} and the latter as \textit{behavioural agreement}.

In this work, we seek to build machine translation (MT) models that satisfy grammatical agreement.
Given an English sentence and a set of attributes (e.g. the gender of the speaker and number of interlocutors), an MT system must translate this sentence into Polish with a correct grammatical agreement to all attributes but introduce no markings of behavioural agreement.

We explore the agreement to one \textsc{Speaker} attribute: the gender of the speaker (\textsc{SpGender}), and three \textsc{Interlocutor} attributes: the gender(s) and number of interlocutor(s) (\textsc{IlGender}, \textsc{IlNumber}), as well as the desired \textsc{formality} of addressing the interlocutor(s). Figure~\ref{fig:example_coh} exemplifies the extent of ambiguity these attributes introduce in English-to-Polish translation.

The \textbf{main contributions} of our work are: (1) a novel English-Polish parallel corpus of TV dialogue annotated for \textsc{SpGender}, \textsc{IlGender}, \textsc{IlNumber} and \textsc{formality}; (2) a tool for analysing attributes expressed in Polish utterances; (3) the examination of a wide range of approaches to attribute control in NMT, showing that at least four of them can be reliably used for incorporating extra-linguistic information within English-to-Polish translation of dialogue.

The paper is structured as follows. Section \ref{sec:related} discusses previous work. Section \ref{sec:problem} presents the problem definition, focusing on Polish as the target language. The creation of the parallel English-Polish corpus of dialogue utterances that mark subsets of the investigated attributes is presented in Section \ref{sec:extraction}. How the MT models are trained to control the four extra-textual attributes is discussed in Section \ref{ssec:training_details}, whilst the results are presented in Section \ref{sec:results}. Finally, we describe conclusions and potential directions for future work in Section \ref{sec:future_work}.

\section{Related Work} \label{sec:related}
The state-of-the-art in MT is currently represented by neural MT (NMT)~\cite{Sutskever2014,Bahdanau2015} implemented via the Transformer architecture~\cite{Vaswani2017}. Despite their unparalleled performance, these models are limited by ignoring the extra-textual context (e.g. speaker's gender). Consequently, much recent work aims to control NMT with various attributes.
In particular, attention has been paid to tasks such as multilingual NMT~\cite{Johnson2017}, by specifying the target language in the input; formality or politeness transfer (e.g.~\newcite{Sennrich2016a}); controlling the gender of the speaker and/or interlocutor~\cite{Elaraby2018,Vanmassenhove2020,moryossef-etal-2019-filling}; length and verbosity~\cite{Lakew2019,Lakew2021}; or constraining the vocabulary~\cite{ailem-etal-2021-encouraging}.

Attribute control in NMT is most commonly facilitated with a \textit{tagging} (or \textit{side constraints}) approach, whereby a set of terms is added to the vocabulary, each embedding a certain type. These are trained alongside token embeddings and used in various ways during inference. Controlling multiple attributes with this approach has not been excessively studied~\cite{schioppa-etal-2021-controlling}, but can be facilitated by simply concatenating the tags~\cite{Takeno2017}. 
However, for a set of equally important attributes, their ordering should not matter, but a tagging approach by design requires tags to be ordered in a specific way. Combining attributes by averaging their embeddings has also been explored in previous work (cf.~\newcite{Lample2019}, \newcite{schioppa-etal-2021-controlling}), where authors incorporated the resulting vectors either into the input of the Encoder or the Decoder or directly into the model~\cite{michel-neubig-2018-extreme,schioppa-etal-2021-controlling}.

Typically, attribute-controlling neural models are fully supervised, requiring annotated training data. Such annotations can be obtained directly, e.g. from metadata~\cite{Vanmassenhove2020}; although most available corpora are unannotated. \newcite{Sennrich2016a} and \newcite{Elaraby2018}
automatically annotate the data using morphosyntactic parsers based on rules, validating agreement to the attribute in question in target-side sentences. To verify that the rules capture the attribute completely, a precision/recall score is computed against a manually labelled test set.

\section{Problem Specification} \label{sec:problem}
Recognising the small number of studies within machine translation research on the English-to-Polish language direction, as well as our capacity (thanks to the available parsers and native speakers to validate their performance), we decide to focus the study on this language pair. Polish is a West Slavic language spoken by over $50$M people over the world~\cite{jassem2003}. It uses an expanded version of the Latin alphabet and is characterised by a complex inflectional morphology~\cite{feldstein2001}. It is a grammatical gender language~\cite{koniuszaniec2003gender} meaning all forms dependent on pronouns must agree to their gender and number. It uses a West Slavic system of honorifics \textit{pani, pan, panie, panowie, państwo} (henceforth \textit{Pan+})~\cite{stone1977address}. Being a null-subject language~\cite{sigursson2009null}, it does not require that pronouns signifying the speaker or the interlocutor are explicit, \textbf{unless} they belong to the \textit{Pan+} group~\cite{Keown2003}. 

English lacks a grammatical gender or a system of honorifics, and the pronoun \enquote{you} is used for both plural and singular second person addressees. It is therefore ambiguous w.r.t. some expressions describing the speaker or the interlocutor, which we capture into four attributes, as follows (the attributes are summarised in Table \ref{tab:types}).

\paragraph{\textsc{Speaker} attributes} The gender of all forms dependent on the pronoun \textit{ja} \enquote{I} must match the gender of the speaker \textsc{SpGender} $\in \{feminine, masculine\}$. 
This includes past and future verbal expressions (e.g. \textit{byłam} `I was\textsubscript{fem}' vs. \textit{byłem} `I was\textsubscript{masc}'), adjectives (e.g. \textit{piękna} `pretty\textsubscript{fem}' vs. \textit{piękny} `pretty\textsubscript{masc}') and nouns (e.g. \textit{wariatka} `lunatic\textsubscript{fem}' vs. \textit{wariat} `lunatic\textsubscript{masc}') that describe the speaker.

\paragraph{\textsc{Interlocutor} attributes} All word forms dependent on the pronoun \textit{ty/wy/Pan+} \enquote{you}, including the pronoun itself, must match:
\begin{itemize}[noitemsep, leftmargin=*, topsep=0pt]
    \item the gender of the interlocutor (\textsc{IlGender}); this includes cases analogous to \textsc{SpGender}, extended to e.g. vocatives (e.g. \textit{Ty waria\underline{tko}/\underline{cie}!} `You lunatic\textsubscript{fem/masc}!');
    \item the number of interlocutors (\textsc{IlNumber}); this includes verbs and pronouns in second person;
    \item the formality in addressing the interlocutor (\textsc{Formality})\footnote{While we define formality as binary, it can be more complex e.g. Japanese in \newcite{feely-etal-2019-controlling}. 
    }; this entails using an inflection of the pronoun Pan+ consistent with \textsc{IlGender} and \textsc{IlNumber} where applicable, or using polite forms (e.g. \textit{Proszę wejść.} `Come in.').
\end{itemize}

\begin{table}[h]
\centering
\scalebox{.73}{
\begin{tabular}{@{}lrl@{}}
\toprule
\multicolumn{1}{c}{Attribute} & \multicolumn{1}{c}{Abbreviation} & Type \\ \midrule
\textbf{\textsc{Speaker}} & \multicolumn{1}{l}{} & \multicolumn{1}{l}{} \\
\hspace{2mm}\multirow{2}{*}{\textsc{SpGender}} & \textit{<sp:feminine>} & Feminine speaker \\
 & \textit{<sp:masculine>} & Masculine speaker \\
 \vspace{2mm}
\textbf{\textsc{Interlocutor}} & \multicolumn{1}{l}{} & \multicolumn{1}{l}{} \\
\hspace{2mm}\multirow{3}{*}{\textsc{IlGender}} & \textit{<il:feminine>} & Feminine interlocutor(s) \\
 & \textit{<il:masculine>} & Masculine interlocutor(s) \\
 & \textit{<il:mixed>} & Mixed-gender interlocutor(s) \\
 \addlinespace[2mm]
\hspace{2mm}\multirow{2}{*}{\textsc{IlNumber}} & \textit{<singular>} & One interlocutor \\
 & \textit{<plural>} & Multiple interlocutors \\
 \addlinespace[2mm]
\hspace{2mm}\multirow{2}{*}{\textsc{Formality}} & \textit{<informal>} & Informal \\
\vspace{2mm}
 & \textit{<formal>} & Formal \\ 
 \bottomrule
\end{tabular}}
\caption{Attributes and types controlled in the experiment.}
\label{tab:types}
\end{table}

Throughout this paper, when discussing \textit{gender} we refer solely to grammatical gender rendered in utterances. In the Polish language, the grammatical system of gender in first and second person is a dichotomy of masculine and feminine variants, lacking alternatives for people who identify as neither. We discuss potential solutions to this issue in directions for future work (\S \ref{sec:future_work}).

\section{Experimental Setup}
\subsection{Data Collection}
\label{sec:extraction}
We collect pre-training data from two corpora: the English-to-Polish part of OpenSubtitles18~\cite{lison-tiedemann-2016-opensubtitles2016} and the Europarl~\cite{koehn2005epc} corpus. The data quantities can be found in Table \ref{tab:data_details} (column \enquote{\texttt{pretrain}}).

\begin{table}[ht]
\centering
\scalebox{.8}{
\begin{tabular}{clccc}
\toprule
\multicolumn{1}{l}{}   &       & \texttt{pretrain} & \texttt{finetune} & \texttt{amb\_test}  \\
\midrule
\multirow{2}{*}{train} & \#sents & $10.8$M & $2.9$M & $-$ \\
                      & \#tokens & $82.1$M  & $26$M & $-$\\
\midrule                       
\multirow{2}{*}{dev}   & \#sents & $3$K & $3.5$K & $-$ \\
                      & \#tokens & $23.3$K & $48.7$K & $-$\\
\midrule
\multirow{2}{*}{test}  & \#sents & $-$ & $3.5$K & $1$K \\
                      & \#tokens & $-$ & $47.7$K & $10.3$K \\
\bottomrule
\end{tabular}}
\caption{Quantities of unique data used for: model pre-training (\texttt{pretrain}), model fine-tuning (\texttt{finetune}) and the test set for calculation of restricted impact (\texttt{amb\_test}). Values are averaged for source and target text.}
\label{tab:data_details}
\end{table}

\begin{figure}[ht]
\centering
\scalebox{1}{
\includegraphics[width=\linewidth]{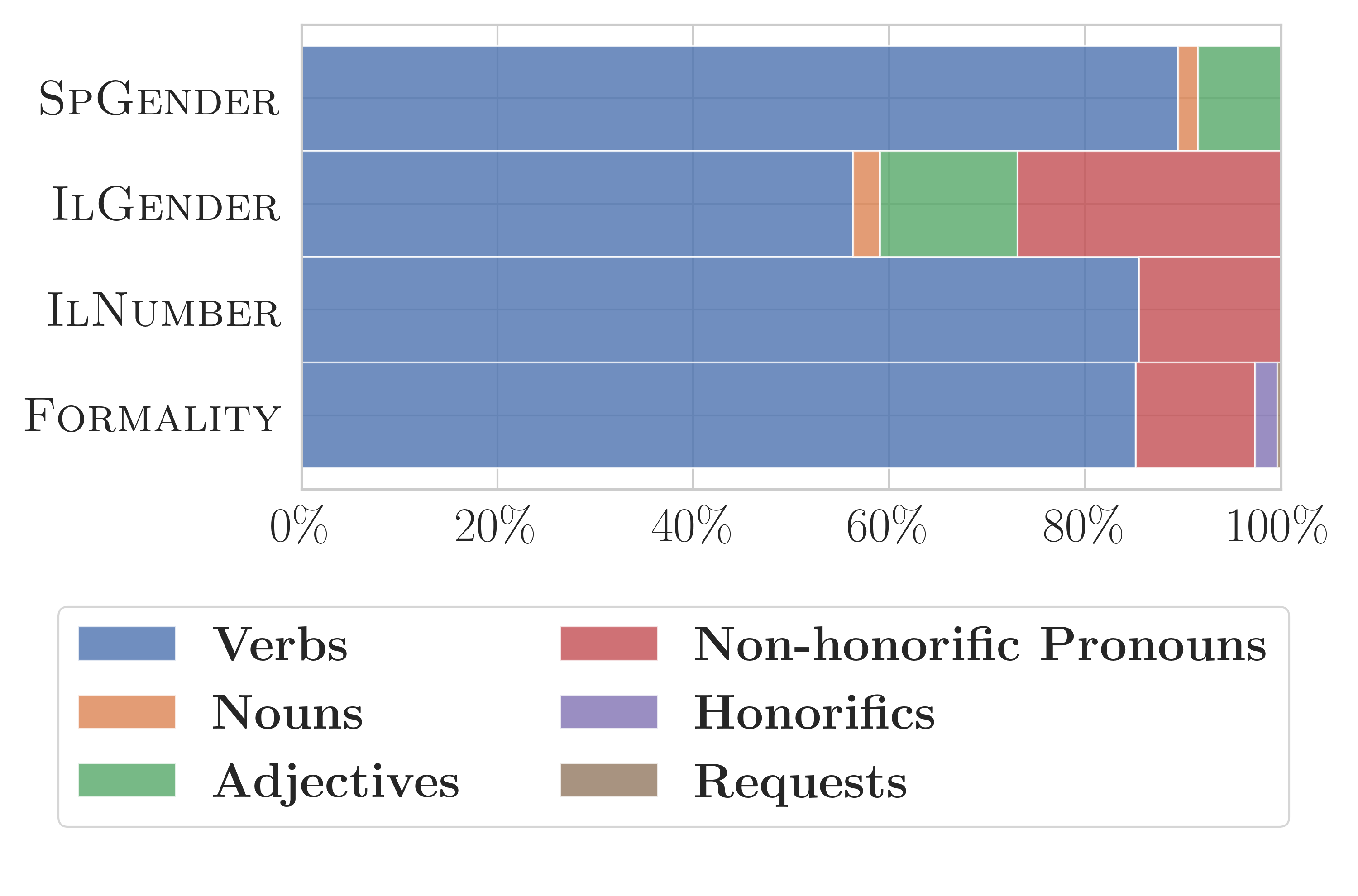}}
\caption{Contributions of each grammatical category to each attribute in the extracted corpus.}
\label{fig:data_quants_adv}
\end{figure}

\paragraph{Corpus Extraction for Fine-tuning}
We extract the fine-tuning data directly from the pre-training corpus; each sample is paired with an annotation of up to four types of attributes. For that purpose we implement a set of morphosyntactic rules for the Polish SpaCy model~\cite{tuo2019spacy} which uses the Morfeusz2 morphological analyser~\cite{kieras2017morfeusz}.\footnote{The code is available at \url{https://github.com/st-vincent1/grammatical_agreement_eamt/.}} 
Since attribute annotations vary at sentence level, we produce sentence-level annotations (instead of word- or scene-level). For both speaker and interlocutor gender attributes, the masculine gender makes up over $60\%$ of the corpus. Altogether, a total of $34.33\%$ of the corpus marks at least one of the attributes.
Figure \ref{fig:data_quants_adv} shows how linguistic categories contributed to extracting each attribute. 

\begin{table*}[h]
\centering
\scalebox{.75}{
\begin{tabular}{@{}cccccccll@{}}
\toprule
\multicolumn{3}{c}{Count} & \multicolumn{4}{c}{Context} & \multicolumn{2}{c}{Example} \\ \midrule
\texttt{train} & \texttt{dev} & \texttt{test} & \textsc{SpGender} & \textsc{IlGender} & \textsc{IlNumber} & \textsc{Formality} & \multicolumn{1}{c}{\textit{English}} & \multicolumn{1}{c}{\textit{Polish}} \\
 \midrule
$419.9$K & $0.8$K & $0.8$K & \textit{sp:feminine} & $*$ & $*$ & $*$ & I'm an amateur. & Jestem amator\textbf{ką}. \\
$743.6$K & $0.8$K & $0.8$K & \textit{sp:masculine} & $*$ & $*$ & $*$ & I'm all alone. & Jestem całkiem sa\textbf{m}. \\
$9.3$K & $0.2$K & $0.2$K & $*$ & \textit{il:feminine} & \textit{plural} & \textit{informal} & You're smitten. & Jeste\textbf{ście} odurz\textbf{one}. \\
$73.8$K & $0.2$K & $0.2$K & $*$ & \textit{il:masculine} & \textit{plural} & \textit{informal} & Have you met Pete? & Pozna\textbf{liście} Pete'a? \\
$315.9$K & $0.2$K & $0.2$K & $*$ & $\times$ & \textit{plural} & \textit{informal} & You need to leave. & Musi\textbf{cie} wyjść. \\
$326.8$K & $0.2$K & $0.2$K & $*$ & $\times$ & \textit{singular} & \textit{informal} & I got you something. & Przyniosłem \textbf{ci} coś. \\
$273.0$K & $0.2$K & $0.2$K & $*$ & \textit{il:feminine} & \textit{singular} & \textit{informal} & Are you sick? & Jeste\textbf{ś} cho\textbf{ra}? \\
$498.7$K & $0.2$K & $0.2$K & $*$ & \textit{il:masculine} & \textit{singular} & \textit{informal} & Understand? & Zrozumia\textbf{łeś}? \\
$0.7$K & $0.1$K & $0.1$K & $*$ & \textit{il:feminine} & \textit{plural} & \textit{formal} & Please, let me explain. & Wyjaśnię \textbf{paniom}. \\
$2.7$K & $0.2$K & $0.2$K & $*$ & \textit{il:masculine} & \textit{plural} & \textit{formal} & Aren't you? & \textbf{Panowie} nie są? \\
$5.7$K & $0.2$K & $0.2$K & $*$ & \textit{il:mixed} & \textit{plural} & \textit{formal} & You are wrong. & My\textbf{lą} się \textbf{państwo}. \\
$63.0$K & $0.2$K & $0.2$K & $*$ & \textit{il:feminine} & \textit{singular} & \textit{formal} & Martini for you? & Dla \textbf{pani} martini? \\
$144.0$K & $0.2$K & $0.2$K & $*$ & \textit{il:masculine} & \textit{singular} & \textit{formal} & Let me have your coat. & Wezmę \textbf{pański} płaszcz. \\
$33.5$K & $0.2$K & $0.2$K & $*$ & $\times$ & $\times$ & \textit{formal} & Go ahead. & \textbf{Proszę} kontynu\textbf{ować}. \\ \bottomrule
\end{tabular}}
\caption{Training data quantities for all combinations of contexts with examples for each combination, with relevant grammatical expressions highlighted. Since \textsc{Speaker} and \textsc{Interlocutor} contexts are always independent, the counts include cases where they co-occur. $* =$ this attribute \textit{may} occur in this place; $\times$ $=$ this attribute is never expressed within this category.}
\label{tab:training}
\end{table*}

Similarly to \newcite{Elaraby2018} and \newcite{gonen-webster-2020-automatically}, we observe that certain nouns marked as describing the speaker or interlocutor have a fixed gender irrespective of that person's gender and are therefore inadequate determinants of their gender (e.g. \textit{coward} \enquote{tchórz} is always masculine).
We could not find a reliable (complete nor heuristic) method to resolve this other than creating a \enquote{stopwords} list of all inflexible nouns. The process is now performed in two steps: we first extract a list of sentences containing gender-marked words and then filter out those that were selected based on our \enquote{stopwords} list of inflexible nouns.

We extract $223.0$K noun-dependent sentences with $9$K unique lemmatised nouns in the first pass, build the \enquote{stopwords} list of $6.8$K words and end up with $67.3$K sentences.

\paragraph{Parser Rules} We identify sentences marking for \textsc{SpGender} by finding tokens in first person singular and verifying that their head marks feminine or masculine gender. \textsc{Formality} is identified through the use of the inflected pronouns in the \textit{Pan+} set (unless it is used as a title, e.g. in \textit{`Ms Smith'}). Formal requests are selected by finding \textit{proszę} (\textit{`please'}) in the target sentence but not in the source. \textsc{IlGender} is trivially inferred in formal cases; for informal language, we match structures analogous to those for the \textsc{SpGender} and extend them to comparative phrases and vocatives. \textsc{IlNumber} follows from the plurality of second-person verbs as well as the use of the pronoun \textit{ty} (\textit{`you'}, singular) or \textit{wy} (\textit{`you'}, plural). 

\paragraph{Parser Performance} To measure the effectiveness of the parser, a native Polish speaker with expertise in NLP manually annotated a random sample of $1$K sentence pairs from the training corpus for the provided attribute types. Given a sample, the annotator was instructed to identify a type from each attribute, and then highlight a part of the Polish sentence proving its occurrence. Precision and recall scores were measured between the judgements of the parser and the annotator. The parser (hereinafter \textit{Detector}) scored near-perfectly (\textbf{99.82\%} precision and \textbf{99.17\%} recall averaged over all attributes) and proved suitable for the tasks of both extracting the corpus and evaluating attribute controlling. Beyond input errors leading to incorrect parsing, we observed two consistent cases of failure: 
\begin{itemize}[noitemsep, leftmargin=*, topsep=0pt]
    \item when the interlocutor is addressed in plural but is in fact singular (in cases like \enquote{Go\textsubscript{singular} help her. Maybe you [two] will\textsubscript{plural} figure it out together.} the addressee may be interpreted as \textit{plural} instead of \textit{singular} depending on the majority of grammatical matches for each type);
    \item some tag questions (e.g. \enquote{prawda?}) or expressions (e.g. the words \enquote{kimś} (`someone\textsubscript{instr.}'), \enquote{czymś} (`something\textsubscript{instr.}')) are consistently incorrectly analysed for dependencies, which sometimes leads to triggering of incorrect rules.
\end{itemize}

\paragraph{Data Selection and Annotation}
\label{ssec:loading}
Table \ref{tab:training} shows particular groups of contexts, their typical expression, and total count in the corpus.\footnote{Note that \textsc{IlGender}, \textsc{IlNumber}, \textsc{Formality} are co-dependent, since they all concern the same entity (the interlocutor), and thus different combinations of their types lead to different grammatical expressions.}
Similarly to \newcite{Sennrich2016a}, we mask the annotations of half the training samples every epoch at random and give half of the unannotated sentence pairs a random set of attributes. This helps preserve the translation quality of the model's outputs when insufficient context is given.

Our development and test sets are balanced across the $14$ context groups (cf. table \ref{tab:training}). We gather a total of $4$K unique examples for each set. When evaluating each implemented approach, we provide two results: when \textit{complete context} is given, or when an \textit{isolated attribute} type is provided. 
Consider a complete-context test case within the \textsc{IlNumber} group of
\begin{align*}
\textit{<il:feminine>,<plural>,<formal>} \text{ I like you.}
\end{align*}
The input for the isolated attribute is as follows:
\begin{align*}
\textit{<plural>} \text{ I like you.}
\end{align*}

that is, we omit all types but those belonging to the examined attribute. For the \textit{complete context} case we provide the full input. To evaluate each individual type (e.g. \textit{<il:feminine>} or \textit{<formal>}), in the isolated attribute case we gather all development/test cases which match the selected type, with a total count of minimum $200$ examples (for \textit{<il:mixed>}) up to $1200$ (for \textit{<plural>}).

\begin{table*}[h]
\centering
\scalebox{.8}{\begin{tabular}{@{}lccc@{}}
\toprule
Approach & \multicolumn{1}{c}{\begin{tabular}[c]{@{}c@{}}Multi-attribute solution\end{tabular}} & \multicolumn{1}{c}{\begin{tabular}[c]{@{}c@{}}Embedding size\end{tabular}} & Input space occupied \\ \midrule
\textit{Types as Tags} &  &  &  \\
\midrule
\hspace{2mm}\textsc{TagEnc}$^\blacktriangle$~\cite{Sennrich2016a} & & & $n_{types}$ \\
\hspace{2mm}\textsc{TagDec}~\cite{Takeno2017} & $++$ & $n_{types} * d_{model}$ & $n_{types} + 1$ \\
\hspace{2mm}\textsc{TagEncDec}$^\blacktriangle$~\cite{Lakew2021} & & & $2 * n_{types} + 1$ \\
\midrule
\textit{Embedded Types} &  &  &  \\
\midrule

\hspace{2mm}\textsc{EmbPWSum}~\cite{Lakew2021} & & & 0 \\
\hspace{2mm}\textsc{EmbAdd}~\cite{schioppa-etal-2021-controlling} & & & 0 \\
\hspace{2mm}\textsc{EmbEnc} (Ours) & $\frac{\sum types}{n_{types}}$ & $n_{types} * d_{model}$ 
& 1 \\
\hspace{2mm}\textsc{EmbSOS}~\cite{Lample2019} & & & 0 \\
\hspace{2mm}\textsc{EmbEncSOS} (Ours) & & & 1 \\
\midrule
\hspace{2mm}\textsc{OutBias}$^\blacktriangle$~\cite{michel-neubig-2018-extreme} & $\frac{\sum types}{n_{types}}$ & $n_{types} * len_{vocab}$ & 0 \\ \bottomrule
\end{tabular}}
\caption{Comparison of examined approaches. $++ =$ concatenation. $^\blacktriangle =$ Approach originally proposed for single-attribute control and extended by us.}
\label{tab:approaches}
\end{table*}

\subsection{Model Settings}
\label{ssec:model_settings}
We use the Transformer architecture~\cite{Vaswani2017} implemented in PyTorch~\cite{Paszke2019}. 
Similarly to \newcite{Lakew2021}, we test a range of model alterations.

We split them into two categories: Types as Tags (\textsc{Tag*}) and Embedded Types (\textsc{Emb*}). We scale each approach that was originally proposed as a way of controlling a single attribute to a multi-attribute scenario:
for \textsc{Tag*}, we supply multiple tags in a random order, and for \textsc{Emb*} we average the embeddings (see Table \ref{tab:approaches}).

\paragraph{Types as Tags}
We encode each type of each attribute as a special vocabulary token (e.g. ${<}singular{>}$, cf. Table \ref{tab:types}). During fine-tuning, these \textit{tags} are concatenated to the source or target\footnote{During inference, we supply tags by forcibly decoding the relevant type tokens, followed by a ${<}null{>}$ token, before the main decoding step commences.} sentences and trained like other tokens. We use three settings: 
\vspace{-.5em}
\begin{itemize}[noitemsep, leftmargin=*]
    \item \textsc{TagEnc}: appending the tags to the source sentence~\cite{Sennrich2016a}.
    \item \textsc{TagDec}: prepending the tag to the target sentence~\cite{Takeno2017}.
    \item \textsc{TagEncDec}: applying tags to both sentences~\cite{Niu2020a}.
\end{itemize}

\paragraph{Average Embedding}

As an alternative to sequential tagging, embedded types $T$ can be averaged and supplied as a single vector $\overline{E(T)}$~\cite{Lample2019}. We test five settings:
\vspace{-.5em}
\begin{itemize}[noitemsep, leftmargin=*]
    \item \textsc{EmbPWSum}: adding $\overline{E(T)}$ position-wise to each input token~\cite{Lakew2021}.
    \item \textsc{EmbAdd}: adding $\overline{E(T)}$ position-wise to Encoder outputs~\cite{schioppa-etal-2021-controlling}.
    \item \textsc{EmbEnc}: concatenating $\overline{E(T)}$ to the input ~(cf. \newcite{dai-etal-2019-style}, but in our approach the embedding is not trained adversarially).
    \item \textsc{EmbSOS}: replace the start-of-sequence (${<}sos{>}$) token in the Decoder input with $\overline{E(T)}$~\cite{Lample2019}.
    \item \textsc{EmbEncSOS}: as an additional setting, we test combining \textsc{EmbEnc} and \textsc{EmbSOS}.
\end{itemize}

As a special case, we test \textsc{OutBias}: adding a type embedding as a bias on the final layer of the Decoder~\cite{michel-neubig-2018-extreme}. We omit the \textit{black-box injection} method of \newcite{moryossef-etal-2019-filling} due to its inapplicability to \textsc{IlGender} in plural and to \textsc{Formality}. Our baseline is the pre-trained model without attribute information.

\label{sec:results}
\begin{table*}[h]
\centering
\scalebox{.8}{
\begin{tabular}{rcccccccc}
\toprule
& \multicolumn{3}{c}{\textit{isolated attribute}} & &  \multicolumn{4}{c}{\textit{complete context}} \\
\midrule
Model & chrF++$^\uparrow$ & BLEU$^\uparrow$ & \textit{Agree}$^\uparrow$ (\%) & & chrF++$^\uparrow$ & BLEU$^\uparrow$ & \textit{Agree}$^\uparrow$ (\%) & \textsc{AmbID}$^\uparrow$ \\
\midrule
Baseline & $46.60$ & $23.13$ & $74.35$ & & $46.60$ & $23.13$ & $74.35$ & $-$\\
\textsc{TagEnc} & $\textbf{48.95}$ & $\textbf{25.52}$ & $99.03$ & &  $\textbf{52.41}$ & $\textbf{29.16}$ & $\textbf{99.39}$ & $\textbf{95.87}$ \\ 
\textsc{TagDec} & $48.65$ & $\textbf{25.40}$ & $99.21$ & & $50.83$ & $27.65$ & $96.84$ & $93.15$ \\ 
\textsc{TagEncDec}& $48.28$ & $25.26$ & $99.35$ & & $51.01$ & $28.15$ & $\textbf{99.26}$ & $82.66$ \\ 
\textsc{EmbPWSum} & $46.03$ & $22.37$ & $100$ & & $51.90$ & $28.69$ & $97.90$ & $88.67$ \\ 
\textsc{EmbAdd} & $47.45$ & $23.61$ & $\textbf{99.96}$ & & $51.77$ & $28.56$ & $98.24$ & $87.76$ \\ 
\textsc{EmbEnc} & $47.72$ & $24.39$ & $83.42$ & & $52.23$ & $\textbf{28.98}$ & $\textbf{99.30}$ & $\textbf{95.58}$ \\ 
\textsc{EmbSOS} & $48.28$ & $24.90$ & $\textbf{99.91}$ & & $\textbf{52.38}$ & $\textbf{29.09}$ & $98.47$ & $92.07$ \\ 
\textsc{EmbEncSOS} & $48.60$ & $25.08$ & $\textbf{99.87}$ & & $51.94$ & $28.77$ & $98.55$ & $92.37$ \\ 
\textsc{OutBias} & $48.59$ & $24.98$ & $96.71$ & & $49.32$ & $26.11$ & $86.25$ & $94.05$ \\ 
\bottomrule
\end{tabular}}
\caption{Translation performance of all models; \enquote{\textit{isolated attribute}} means that only one (the investigated) attribute was revealed to the model. The highlighted scores include the best one in the column and all statistically equivalent results according to a bootstrap resampling method (p < 0.05).}
\label{tab:agreement_results}
\end{table*}

\subsection{Training Details}
\label{ssec:training_details}
We preprocess the corpus with Moses tools for detokenisation and normalising punctuation\footnote{\url{https://github.com/alvations/sacremoses}}, and by running a short set of custom rules. We train a joint sub-word segmentation model of $16$K tokens with SentencePiece~\cite{kudo-richardson-2018-sentencepiece} and encode both sides of the corpus. We follow the standard training regimen for a $6$-layer Transformer~\cite{Vaswani2017} with an input length limit of $100$ tokens; this model has just over $52.3$M trainable parameters. All training is done on a single $32$GB GPU. As the decoding algorithm, we use beam search with a beam size of $5$. We pre-train the model until a patience criterion of the chrF++~\cite{popovic-2017-chrf} validation score not increasing for $5$ consecutive validation steps (which occur every $3/4$th epoch). This happens around the $24$th epoch, or after $66$ hours of training.

Each of the nine architectural upgrades is a copy of the pre-trained model expanded with the relevant component and fine-tuned. The fine-tuning process exposes the model to the fine-tuning corpus in $10$ epochs; performance is validated every half epoch. We select the best checkpoint based on the highest chrF++ score on the development set.

\subsection{Evaluation}
\label{ssec:eval_template}
We consider the following criteria in evaluation:
\begin{enumerate}[noitemsep, leftmargin=*]
    \item \textbf{Translation Quality.} Attribute-controlled translations should be of quality no worse than translations of the non-specialised model.
    \item \textbf{Grammatical Agreement.} Attribute-controlled hypotheses should completely agree to the specified type where necessary.
    \item \textbf{Restricted Impact.} Grammatical agreement should only affect words that explicitly render the attributes. Therefore, if no attribute is to be expressed in the hypotheses, then they should be no different from baseline hypotheses. 
\end{enumerate}

We evaluate translation quality with chrF++~\cite{popovic-2017-chrf}\footnote{For clarity, we normalise chrF++ scores to a $[0,100]$ range.} and BLEU~\cite{papineni-etal-2002-bleu}. Grammatical agreement is quantified with the help of the \textit{Detector}. For every attribute, we calculate how many hypotheses agree to the correct type $t$ and to the incorrect type $\hat{t}$. Let $hyp_t$ be a hypothesis translated using type $t$ as context, and $agree(hyp, t)$ denote that the \textit{Detector} has found evidence of type $t$ expressed in $hyp$. We express the total agreement score as:
\[\textit{Agree} = \frac{agree(hyp_t, t)}{agree(hyp_t, t) + agree(hyp_t, \hat{t})}\]
Finally, we quantify restricted impact with a custom metric, which measures that attribute-independent sentences do not carry any attribute-reliant artifacts; we define this metric, \textsc{AmbID}, as:
\[\text{chrF++}(\text{NMT}(src_a, A), \text{NMT}(src_a, \widehat{A}))\]
where $A$ is a set of attribute types and $\widehat{A}$ is the reverse set.\footnote{For the type triplet \textsc{IlGender}
we assume that $\widehat{\textit{il:masculine}} = \textit{il:feminine}$, $\widehat{\textit{il:mixed}} = \textit{il:feminine}$, $\widehat{\textit{il:feminine}} = \textit{il:masculine}$.} 
We use an attribute-ambivalent test set of a $1$K sentences to calculate this score (Table \ref{tab:data_details}, column \enquote{\texttt{amb\_test}}).

\section{Results}
\begin{figure*}[ht]
\centering
\scalebox{.99}{
\includegraphics[width=\linewidth]{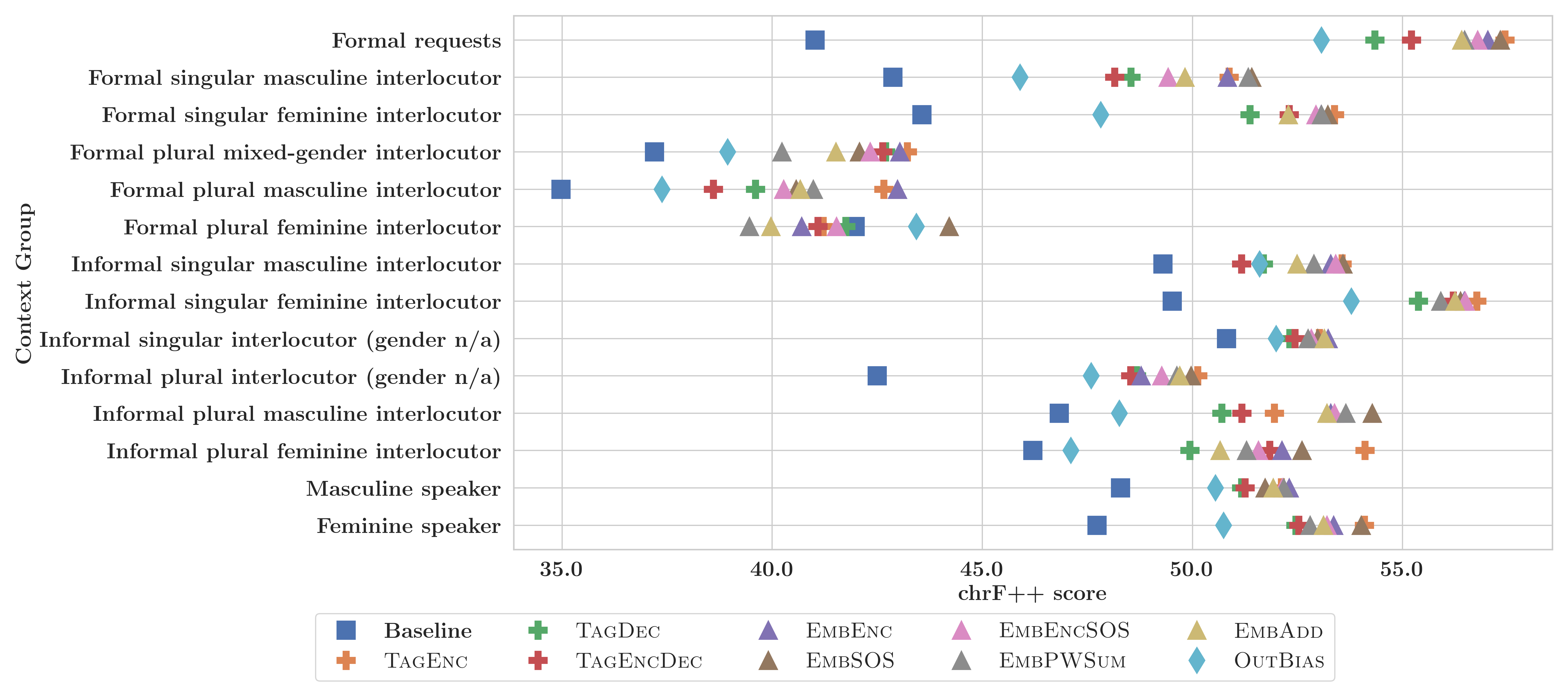}}
\caption{Translation quality (chrF++) for each contextual group.}
\label{fig:chrf_per_group}
\end{figure*}
We report quantitative results in Table \ref{tab:agreement_results}.

\paragraph{Grammatical Agreement}
The \textit{Agree} column in Table \ref{tab:agreement_results} shows the agreement scores given by the \textit{Detector}. 
In the isolated attribute scenario, all methods but \textsc{OutBias} and \textsc{EmbEnc} achieve near-perfect agreement scores. The agreement scores in the \textit{complete context} scenario remain high for other models except \textsc{TagDec}, and pick up for \textsc{EmbEnc}, suggesting that controlling several attributes generally has no negative impact on individual attributes.

\paragraph{Translation Quality}
Attribute-controlling models achieve significant gains over baseline for both the isolated attribute and complete context scenarios, and the gains are consistently higher in the latter, suggesting that exposing the models to more context yields better translations. \textsc{TagEnc} achieves the highest improvement over the baseline in terms of chrF++/BLEU for complete context ($+5.81 \text{ chrF++}/{+}6.03 \text{ BLEU}$). The gains in translation quality are correlated with agreement scores, except for \textsc{EmbPWSum}, for which the isolated attribute scenario leads to a near-perfect agreement but low quality scores. Further investigation shows that this model learned to overproduce context-sensitive words when given a context of only a subset of types (e.g. translating \enquote{you} as \enquote{I} to introduce \textsc{SpGender} marking), leading to high agreement scores but degradation in quality. This highlights the importance of pairing an accuracy measure with a translation quality metric.

To investigate how successful the models are at modelling each context group individually, we report the mean chrF++ scores obtained for each group's test set (Figure \ref{fig:chrf_per_group}). All contextual models bring significant improvements over the baseline except in the \textit{Formal plural feminine interlocutor group}, for which there was little training data (cf. Table \ref{tab:training}); improvements are consistently greater for feminine than masculine groups. No single model performs consistently better than others than others, but \textsc{TagDec}, \textsc{EmbPWSum} and \textsc{OutBias} fall behind on most groups. Finally, we observe no significant gain generally from including information in both the Encoder and the Decoder.

\paragraph{Restricted Impact}
The \textsc{AmbID} scores shown in Table \ref{tab:agreement_results} reveal that \textsc{TagEnc} and \textsc{EmbEnc} introduce the least variation in attribute-ambivalent utterances, suggesting that adding contextual information to the Encoder input only helps limit creation of unwanted artifacts. The distance of only $4.13$ chrF++ points to the ideal score of $100$ for the highest-scoring model suggests good separation of grammatical and behavioural agreement. Some separation-specific modelling may further improve this score, but it was outside the scope of this work.

\paragraph{General Discussion}

The results suggest that \textsc{TagEnc} is the most reliable approach to the presented problem, followed by \textsc{EmbSOS} and \textsc{EmbEnc}. Notably, we find other methods dubbed as superior to \textsc{TagEnc} in previous work (\textsc{EmbAdd}, \textsc{TagDec} and \textsc{TagEncDec}) to underperform in our case. 

\section{Conclusions and Future Work}
\label{sec:conclusions}
\label{sec:future_work}
In this work, we have highlighted the problem of grammatical agreement in translation of TV dialogue in the English-to-Polish language direction. We have created and described a dataset annotated for four speaker and interlocutor attributes that directly influence grammar in dialogue: speaker's gender, interlocutor's gender and number and formality relations between them. We have presented a selection of models capable of controlling these attributes in translation, yielding a performance gain of up to ${+}5.81$chrF++/${+}6.03$BLEU over the baseline (non-controlling) model. Finally, we have produced a tool that produces an accuracy score for agreement to each type.

Considering all criteria of evaluation, we have identified \textsc{TagEnc} as the best performing approach, with \textsc{EmbEnc}, and \textsc{EmbSOS} also achieving competitive performance. \textsc{TagEnc} may be more attractive in scenarios where interventions in the model architecture are impossible as it can be implemented via data preprocessing alone, but the other two have a more scalable design (cf. \S \ref{sec:related}). Finally, contrary to some previous work, we found no advantages stemming from including the contextual information in the Decoder as well as the Encoder.

\paragraph{Future Work} 
NMT research should strive to move beyond seeing gender as a dichotomous phenomenon~\cite{savoldi-etal-2021-gender}. Within this paper we did not consider the scenarios with non-binary interlocutors due to i) lack of available data and ii) lack of consensus regarding non-binary gender expression in the Polish language~\cite{Misiek2020MisgenderedIT}. However, our work can be applied to non-binary expression once data and more studies are available. Furthermore, the influence in NMT of other extra-textual attributes (e.g. multimodal ones, like spatial information, or emergent ones, such as personal attributes) is yet to be explored. It remains an open question whether such attributes should all be considered individually, or whether there is a way of identifying and/or using them implicitly.

\section*{Acknowledgements}
This work was supported by the Centre for Doctoral Training in Speech and Language Technologies (SLT) and their Applications funded by UK Research and Innovation [grant number EP/S023062/1].

\bibliographystyle{eamt22}
\bibliography{anthology,references}

\end{document}